\documentclass[runningheads]{llncs}
\usepackage{graphicx}
\usepackage{graphicx}

\usepackage{tabularx}
\usepackage{algorithm}
\usepackage{watermark}
\usepackage{hyperref}

\usepackage{verbatim}
\usepackage{array}
\usepackage{multirow}
\usepackage{pifont}
\usepackage{amssymb}
\usepackage{mathtext}
\usepackage{booktabs,siunitx} 
\usepackage{mathtools}
\usepackage{fontenc}
\usepackage{microtype}
\usepackage{xcolor}
\usepackage{colortbl}
\usepackage{textcomp}
\usepackage{url}
\usepackage{subfigure}
\usepackage{appendix}

\usepackage{enumerate}

\usepackage{algorithm}
\usepackage{array}
\usepackage{eqparbox}
\usepackage{array}
\usepackage{eqparbox}
\usepackage{flushend}

\usepackage[normalem]{ulem}
\usepackage{algpseudocode}
\usepackage{float}
\floatstyle{plaintop}
\restylefloat{table}
\definecolor{sdcor}{RGB}{0,0,0}
\definecolor{modreq}{RGB}{0,0,0}
\definecolor{sdrm}{RGB}{0,0,0}

\begin{document}
\title{Confidence Score for Unsupervised Foreground Background Separation of Document Images}
%
%
\author{Soumyadeep Dey \and Pratik Jawanpuria}
\authorrunning{S. Dey and P. Jawanpuria}
%
\institute{Microsoft, India Development Center \\
\email{\{Soumyadeep.Dey,Pratik.Jawanpuria\}@microsoft.com}}
\maketitle              
\begin{abstract}
Foreground-background separation is an important problem in document image analysis. Popular unsupervised binarization methods (such as the Sauvola's algorithm) employ adaptive thresholding to classify pixels as foreground or background. In this work, we propose a novel approach for computing confidence scores of the classification in such algorithms. This score provides an insight of the confidence level of the prediction. The computational complexity of the proposed approach is the same as the underlying binarization algorithm. Our experiments illustrate the utility of the proposed scores in various applications like document binarization, document image cleanup, and texture addition. 


\keywords{Binarization \and Cleanup \and Confidence score.}
\end{abstract}
\section{Introduction}
\label{sec:intro}

The technique to classify foreground and the background pixels is known as binarization.  
Various supervised and unsupervised techniques have been reported in literature for document image binarization.
The simplest method to achieve binarization is thresholding gray-scale or color document images.
Analytical techniques for document image binarization involve segmenting the foreground pixels and background pixels based on some threshold. 
For binarization of an image, a global threshold is computed based on the distribution of pixel intensities in~\cite{otsu}. 
Sauvola and Pietikäinen proposed an adaptive thresholding method for document image binarization in~\cite{Sauvola00adaptivedocument}.  
In this method, a threshold is computed for each pixel based on local mean and variance surrounding the pixel. 
Lazzara and Geraud proposed proposed a multi-scale version  of Sauvola's algorithm in~\cite{lazzara_IJDAR14} to make it adaptable for low contrast images. 
The above techniques use pixel intensity based information to perform local/global threshold to obtain binary image. 
In contrast, Peng~\textit{et~al.}~\cite{peng_binary_icdar13} proposes a Gabor filter based stroke orientation computation technique for document binarization task.
A fast Fuzzy C-Means clustering based document binarization technique has been proposed in~\cite{binary_fcmean_icdar19}. 
A regression based method for background estimation is proposed in~\cite{Vo_bin_PR18}. The estimated background is subtracted from the input image and global thresholding is applied for the binarization task. 
In recent time, document binarization is also explored using supervised techniques like maximum entropy classification~\cite{Liu_Binary_ICDAR17},  multi-resolutional attention model~\cite{binary_attention_icdar19},  convolutional neural network~\cite{deyICDAR21_cleanup}.

There are various downstream applications of foreground segmentation from the background pixels. 
However,  mere binary level segmentation is not enough to achieve these downstream tasks,  since perfect segmentation of foreground from background pixels is difficult to achieve and very much data dependent in case of supervised methods.  
Therefore it is important to have a confidence score for each pixels in an unsupervised manner.  
In this work,  we have proposed an unsupervised scoring function for each pixel of an image to define its confidence to be labeled as background or foreground.  
The primary contribution of the paper lies in defining the scoring function in an unsupervised manner.  
We have also shown the application of these scores in various document processing techniques.  


\section{Computation of scores for each pixel}
\label{sec:score}

In the Sauvola's algorithm~\cite{Sauvola00adaptivedocument}, a threshold is computed for each pixel using the Eq~\ref{eq:sauvola}, where, for an input image $I$,  $R=\frac{\textrm{max}(I) - \textrm{min}(I)}{2}$. 
\begin{equation}
T_W (p)= m_{W}^{p} \times [1+k\times (\frac{s_{W}^{p}}{R} - 1)]
\label{eq:sauvola}
\end{equation}
The threshold $T$ is computed for each pixel ($p$) based on a window $W$ of size $n \times n$ surrounding it, where $m_{W}^{p}, s_{W}^{p}$ respectively represent mean and standard deviation of $W$ around pixel $p$, and  $k$ lies between $0 \le k \le 1$. 

\textcolor{sdcor}{Empirically, it has been observed that binarization using the the thresholds obtained from Eq~\ref{eq:sauvola} misses foreground pixels in many scenarios. 
An example failure case is shown in Fig.~\ref{fig:applications}(A)(ii). 
Such a segmented output may also used in downstream applications such as image clean-up task. An example output of such a setting is provided in Fig.~\ref{fig:applications}(B)(ii). We again observe a substantial loss of foreground information with the segmented output obtained using Eq~\ref{eq:sauvola}.}

\textcolor{sdcor}{To alleviate the above concern, we propose to compute a confidence score for each pixel. The goal of this score is to reflect the confidence about the class prediction. Typically, confidence on prediction should increase if the pixel value lies further away from the computed threshold. Based on this intuition, we define normalized confidence values of background ($C^{b}_W(p)$) and foreground ($C^{f}_W(p)$), for each pixel $p$ using Eqs~\ref{eq:bconf} and~\ref{eq:fconf}. }
\begin{equation}
  C^{b}_W(p)  = \left\{
  \begin{array}{l l}
    \frac{I(p) - T_W(p)}{\textrm{max}(I) - T_W(p)} & \quad \textrm{if \textit{$I(p)$} $>$ \textit{$T_W(p)$}}\\
    1 -  \frac{T_W(p)- I(p)}{T_W(p) - \textrm{min}(I)} & \quad  \textrm{otherwise}\\
  \end{array} \right. 
\label{eq:bconf}
\end{equation} 
\begin{equation}
C^{f}_W(p)  = 1 - C^{b}_W(p)
\label{eq:fconf}
\end{equation} 
Here, max($I$) and min($I$) represent maximum and minimum value of any pixel of an input image $I$, respectively. It should be noted that the confidence score lies in the interval $[0,1]$. Overall, with the availability of such scores, downstream tasks may take a more suitable decision (for pixels with low confidence scores) to avoid foreground information loss.  
\textcolor{sdcor}{The proposed confidence scores can be generated with any adaptive thresholding approach. For empirical comparison, we considered Sauvola's thresholding algorithm~\cite{Sauvola00adaptivedocument} as the base method.}

\section{Applications}
\label{sec:applications}

\begin{figure}[ht]
 \centering
 \begin{tabular}{@{}c@{\ }c@{\ }c@{\ }c@{}}
(A) &
\fbox{\includegraphics[width=.28\textwidth]{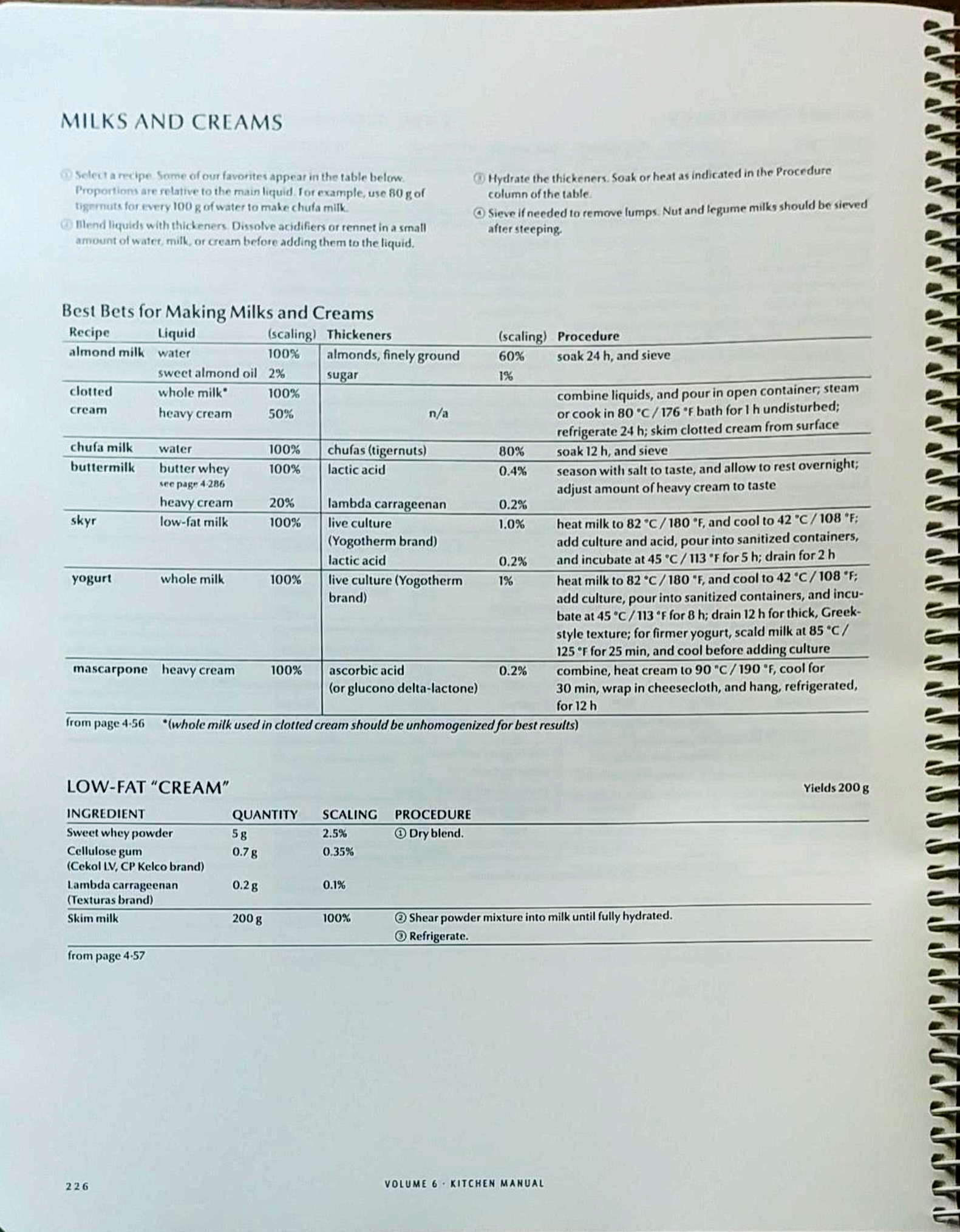}} &
\fbox{\includegraphics[width=.28\textwidth]{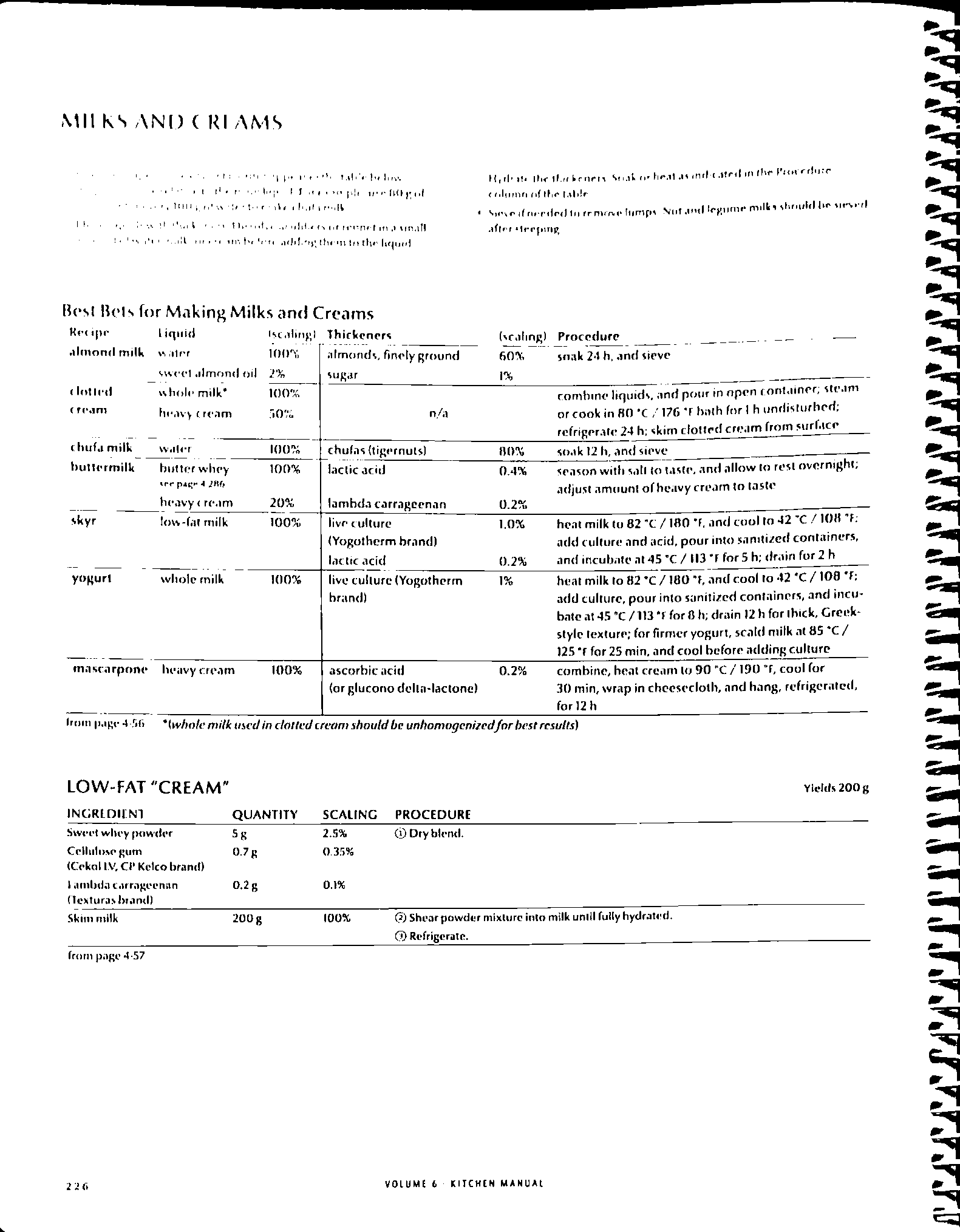}} &
\fbox{\includegraphics[width=.28\textwidth]{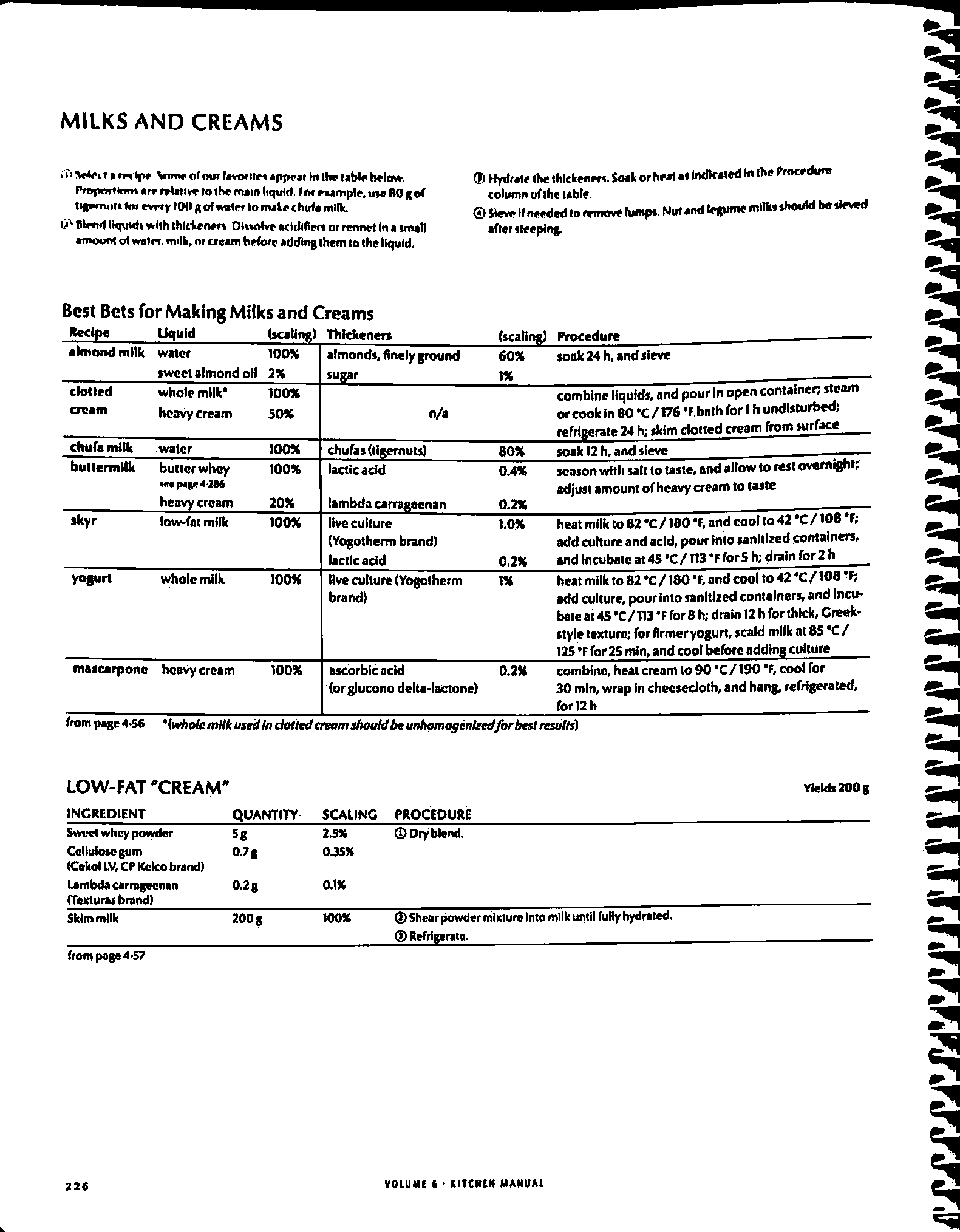}}\\ 
(B) &
\fbox{\includegraphics[width=.28\textwidth]{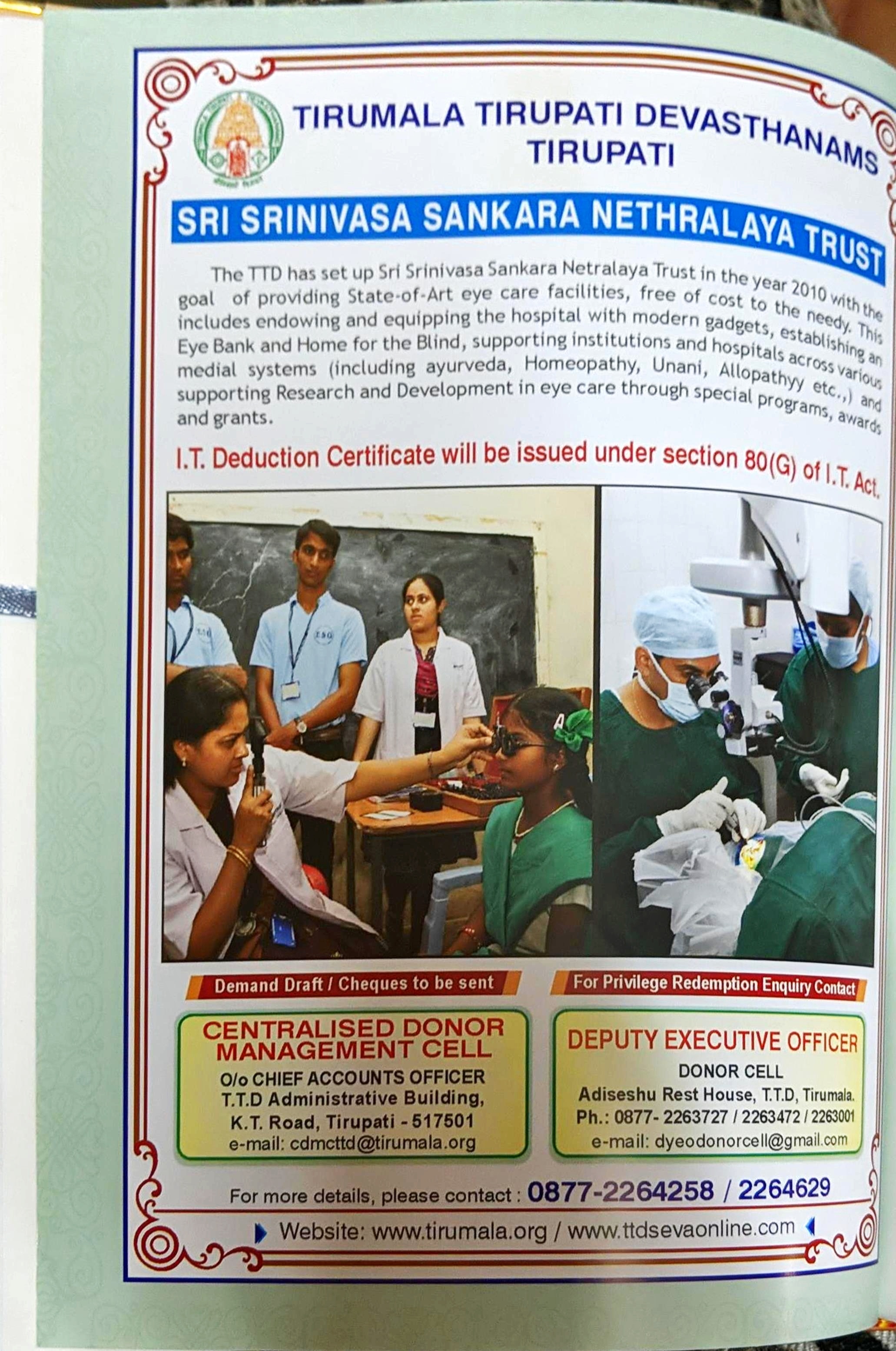}} &
\fbox{\includegraphics[width=.28\textwidth]{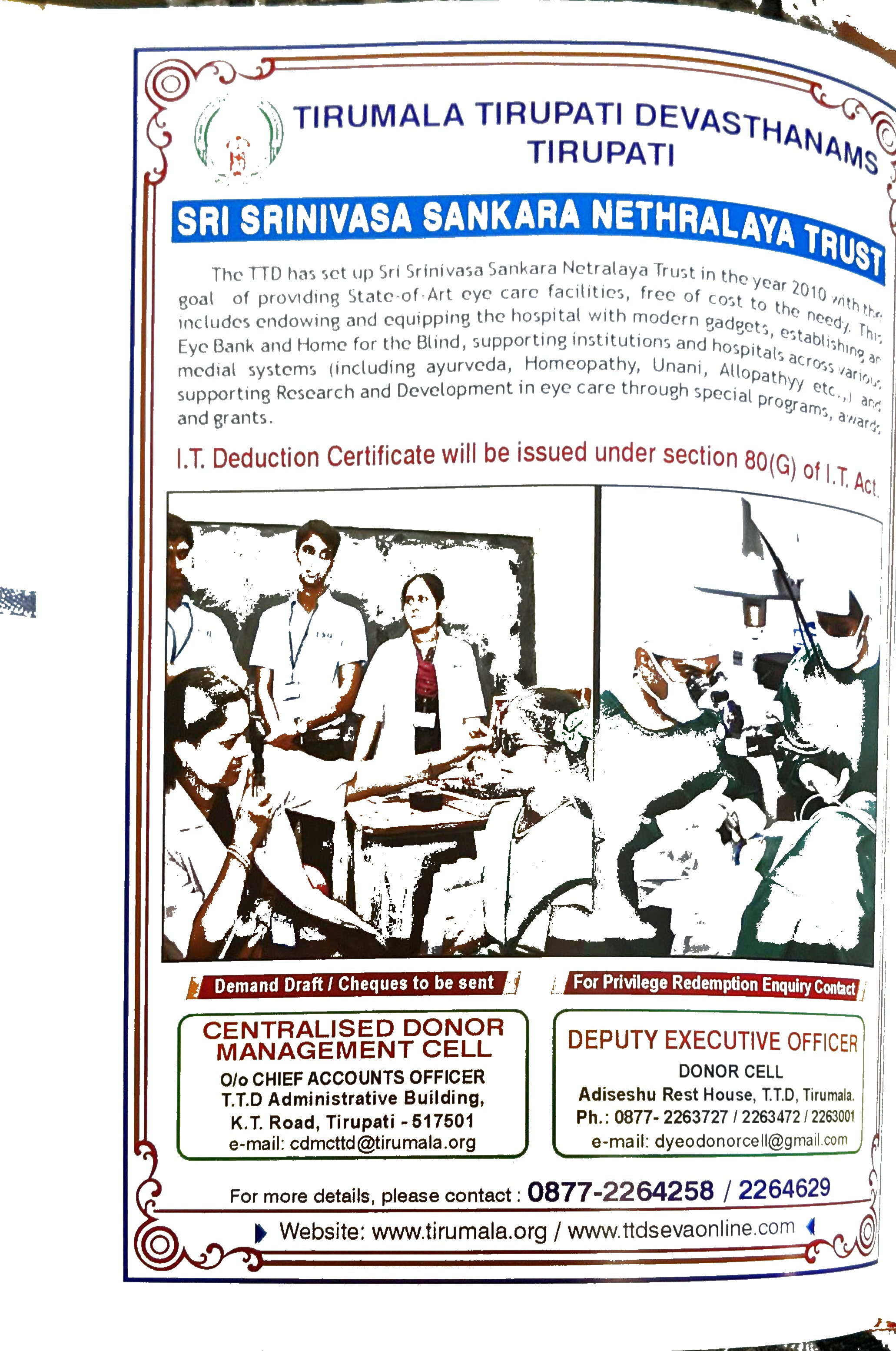}} &
\fbox{\includegraphics[width=.28\textwidth]{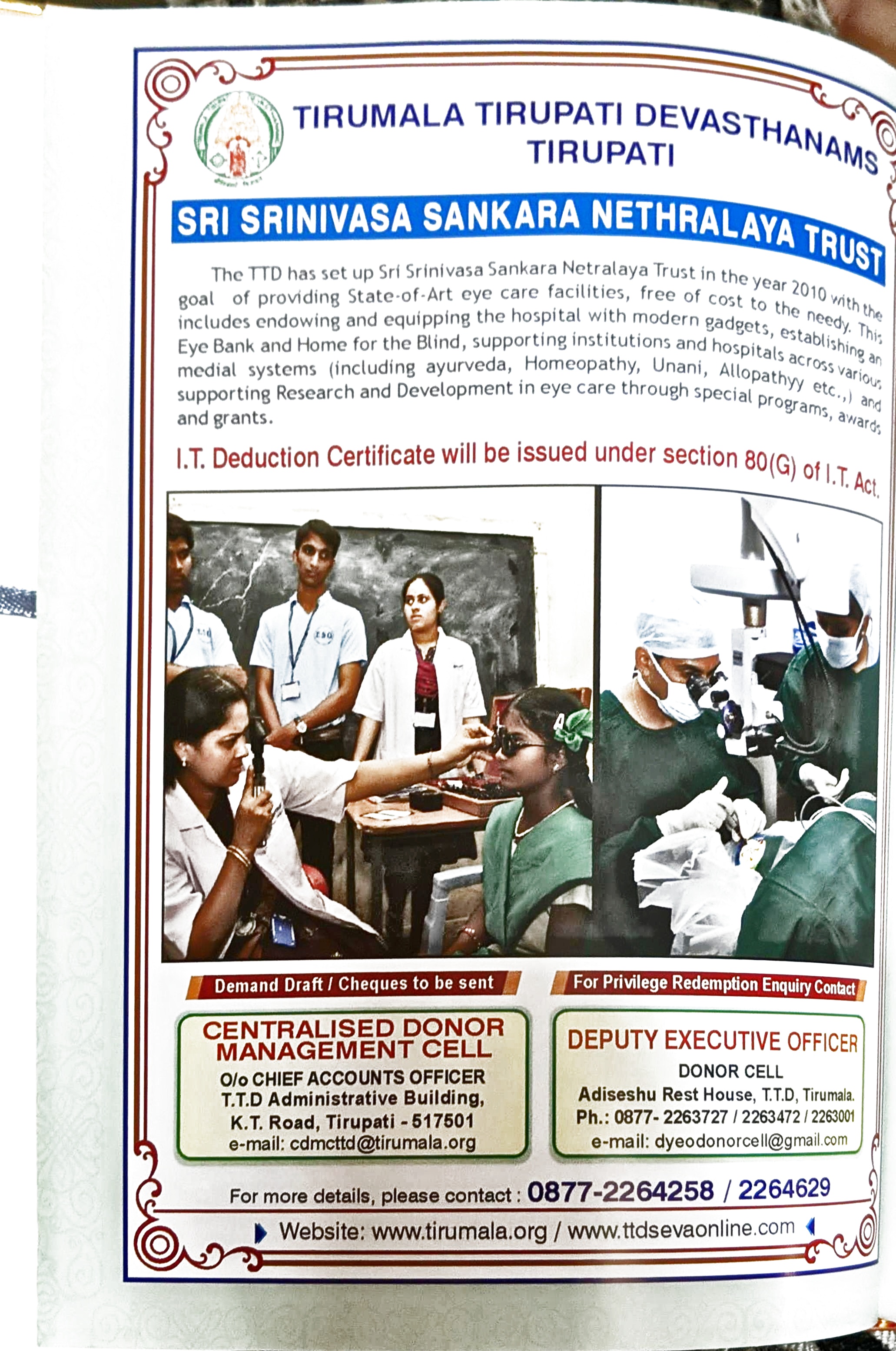}}\\ 
(C) &
\fbox{\includegraphics[width=.28\textwidth]{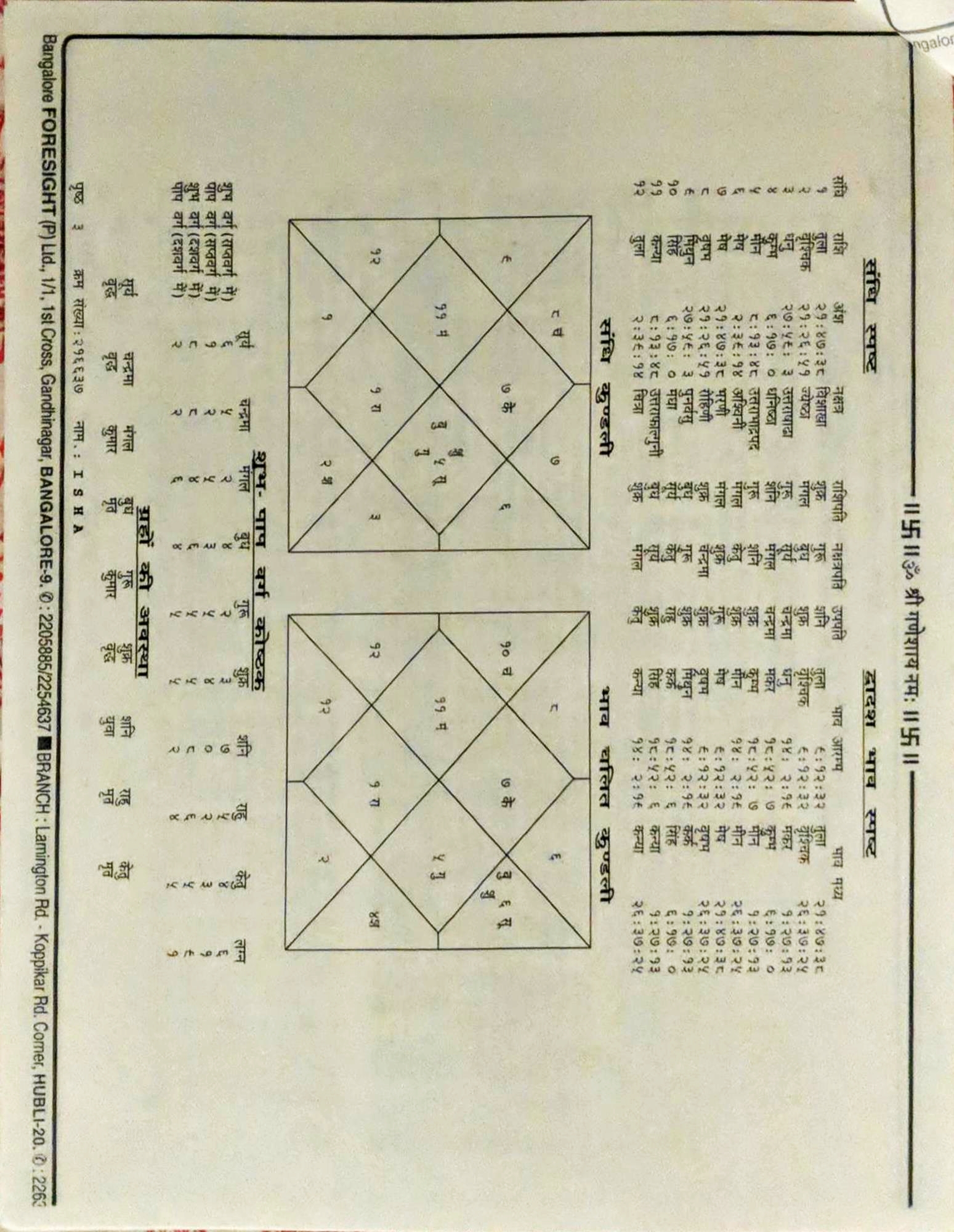}} &
\fbox{\includegraphics[width=.28\textwidth]{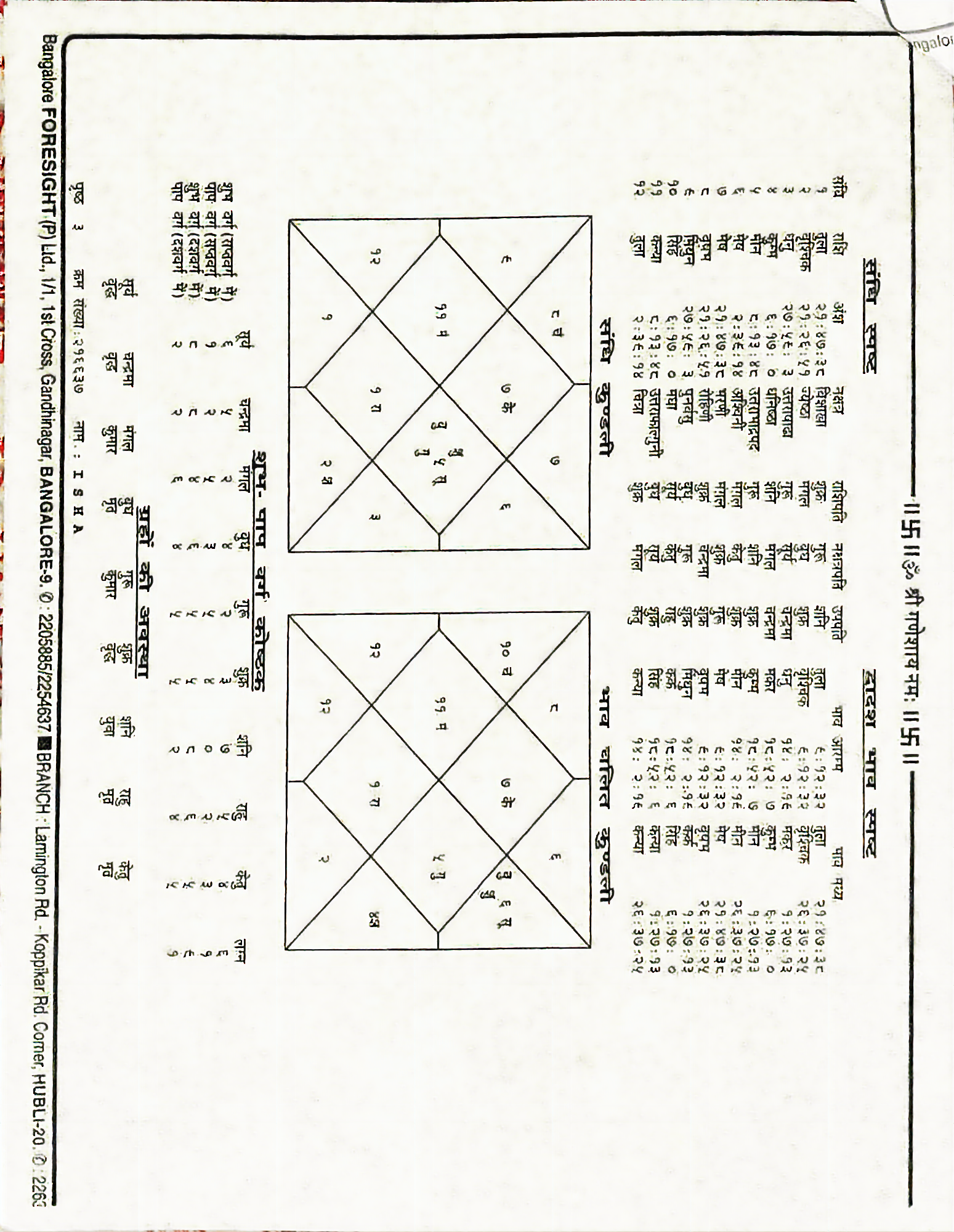}} &
\fbox{\includegraphics[width=.28\textwidth]{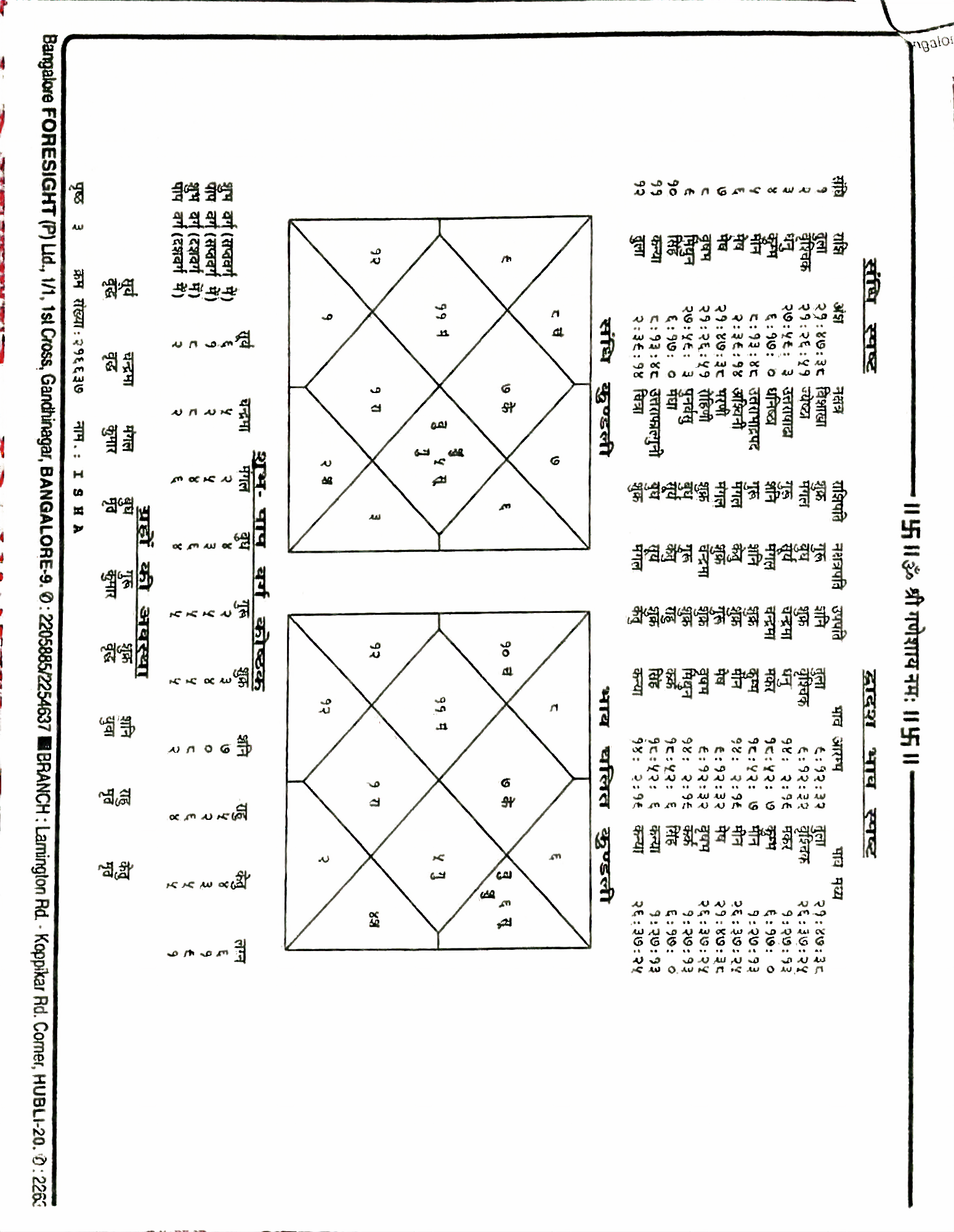}}\\ 
(D) &
\fbox{\includegraphics[width=.28\textwidth]{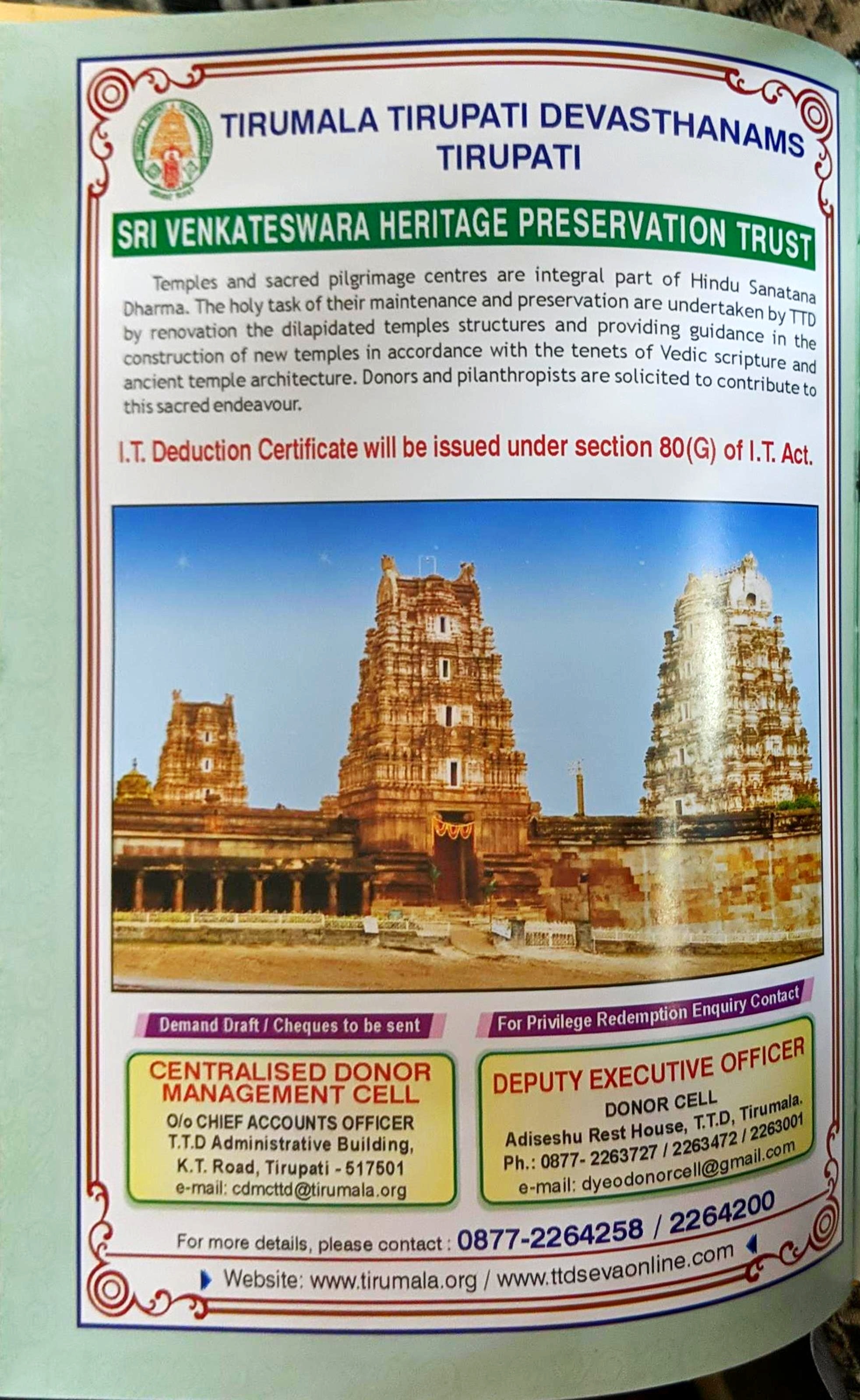}} &
\fbox{\includegraphics[width=.28\textwidth]{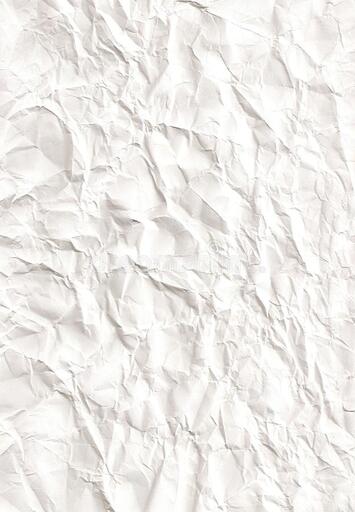}} &
\fbox{\includegraphics[width=.28\textwidth]{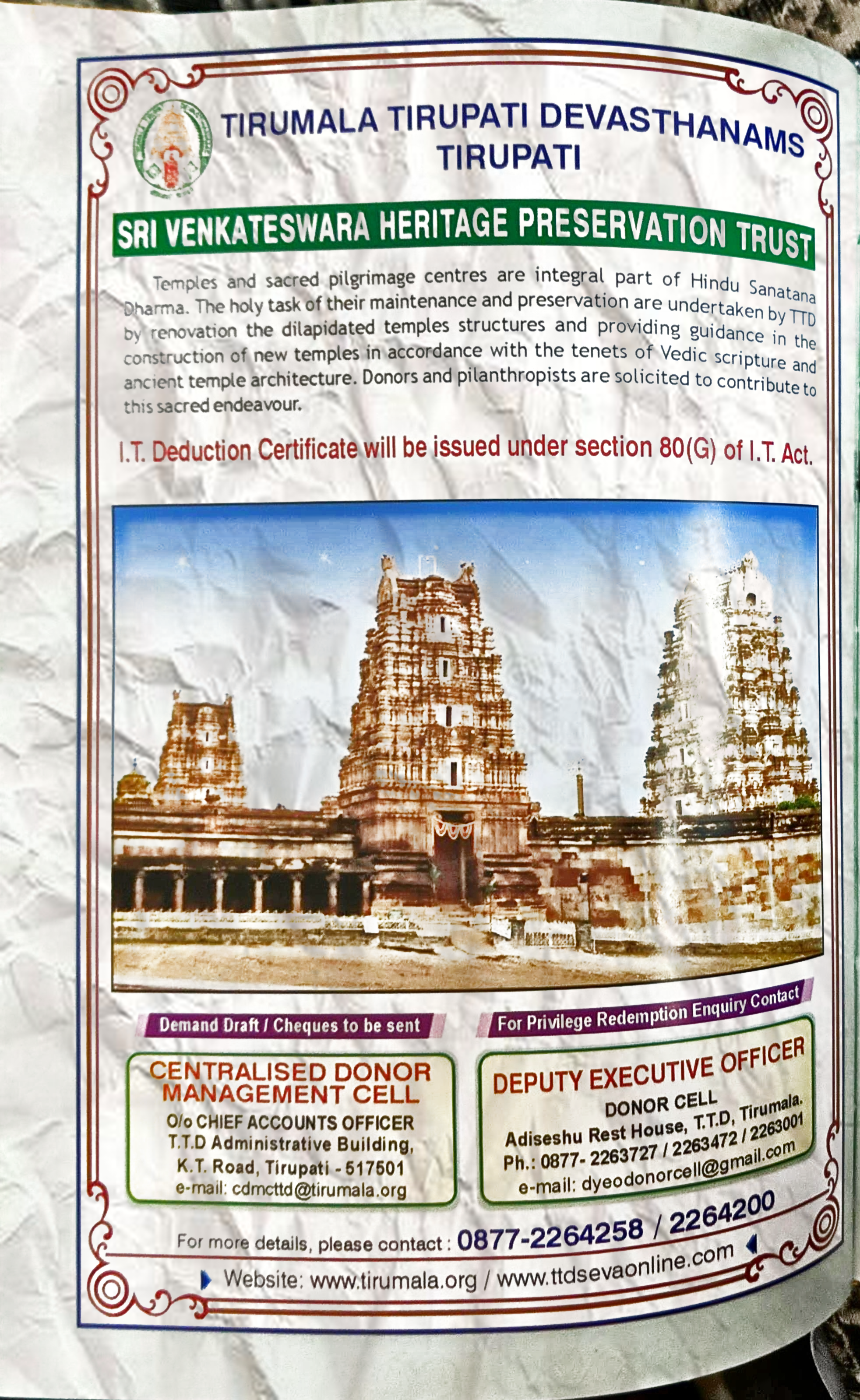}}\\ 
 & (i) & (ii) & (iii) \\
\end{tabular}
\caption{(A,B,C,D)(i)~input images; A(ii)~binary image using~\cite{Sauvola00adaptivedocument}; A(iii)~improved binary image using the proposed scores; B(ii)~image cleanup using~\cite{Sauvola00adaptivedocument}; B(iii)~image cleanup using the proposed scores; C(ii)~image cleanup using~\cite{deyICDAR21_cleanup}; C(iii)~image cleanup using~\cite{deyICDAR21_cleanup} with proposed score based pre-processing; D(ii)~texture to be transferred to D(i); D(iii)~texture transferred image using proposed scores.}
\label{fig:applications}
\end{figure}

We discuss four applications of the proposed score values. \\
\textbf{Document binarization}: We develop a modified version of~\cite{Sauvola00adaptivedocument} to handle missing data using the pixel scores. Our code is available at  \url{https://tinyurl.com/scoredbinarization}. Our result for a sample image is in Fig.~\ref{fig:applications}(A)(iii).     \\
\textbf{Document cleanup}: The missing data from~\cite{Sauvola00adaptivedocument} results in patchy cleanup. 
The proposed score function can be used to obtain a non-patchy cleaned up version of the input image (please refer to Fig.~\ref{fig:applications}(B)).  \\
\textbf{Preprocessing}: The proposed score values can also be used as pre-processing step to various algorithms. For instance, its application as preprocessing step to DNN based image cleanup~\cite{deyICDAR21_cleanup} is discussed in Fig.~\ref{fig:applications}(C). \\
\textbf{Texture transfer}: The goal here is to transfer the content of an input image to a new textured background without any loss of original content of the original document.  
Here, we used the foreground and background scores to achieve this objective. An example of such texture transfer is shown in Fig.~\ref{fig:applications}(D).

\section{Conclusion}
\label{sec:conclusion}

We presented an unsupervised method to compute confidence score for each pixel in a document image. 
We have further shown the utilization of these computed scores for various downstream tasks. 
More theoretical investigation and experimentation on the discussed applications are interesting future directions.

%
%
%
%

\end{document}